\documentclass[conference]{IEEEtran}
\IEEEoverridecommandlockouts
\usepackage{cite}
\usepackage{amsmath,amssymb,amsfonts}
\usepackage{algorithmic}
\usepackage{graphicx}
\usepackage{textcomp}
\usepackage{xcolor}

\usepackage{url}
\usepackage{booktabs}
\usepackage{graphicx}

\def\BibTeX{{\rm B\kern-.05em{\sc i\kern-.025em b}\kern-.08em
    T\kern-.1667em\lower.7ex\hbox{E}\kern-.125emX}}
\begin{document}

\title{Disentangling Shared and Task-Specific Representations from Multi-Modal Clinical Data\\
\thanks{This work has been funded by the National Natural Science Foundation of China (82471535 to H.S.). The first two authors contributed equally to this work, and the last two authors contributed equally as co-senior authors.}
}

\DeclareRobustCommand*{\IEEEauthorrefmark}[1]{%
    \raisebox{0pt}[0pt][0pt]{\textsuperscript{\footnotesize #1}}%
}

\author{
    \IEEEauthorblockN{
        He Lyu\IEEEauthorrefmark{1,2,3},
        Huolin Zeng\IEEEauthorrefmark{1,2},
        Junren Wang\IEEEauthorrefmark{1,2},
        Huazhen Yang\IEEEauthorrefmark{1,2},
        Linchao He\IEEEauthorrefmark{4},
        Yong Chen\IEEEauthorrefmark{1,2}, \\
        Zhirui Li\IEEEauthorrefmark{5},
        Andreas Maier\IEEEauthorrefmark{3},
        Siming Bayer\IEEEauthorrefmark{3} and
        Huan Song\IEEEauthorrefmark{1,2,6,7 *}
    }
    \vspace{1.0 ex}
    \IEEEauthorblockA{
        \IEEEauthorrefmark{1}Department of Anesthesiology and West China Biomedical Big Data Center\\
        West China Hospital, Sichuan University, Chengdu, China\\
        \IEEEauthorrefmark{2}Med-X Center for Informatics, Sichuan University, Chengdu, China\\
        \IEEEauthorrefmark{3}Pattern Recognition Lab, Friedrich-Alexander-Universität Erlangen-Nürnberg, Erlangen, Germany\\
        \IEEEauthorrefmark{4}Department of National Key Laboratory of Fundamental Science on Synthetic Vision, Sichuan University, Chengdu, China\\
        \IEEEauthorrefmark{5}Sichuan University, Chengdu, China\\
        \IEEEauthorrefmark{6}Institute of Environmental Medicine, Karolinska Institutet, Stockholm, Sweden\\
        \IEEEauthorrefmark{7}Center of Public Health Sciences, Faculty of Medicine, University of Iceland, Reykjavík, Iceland\\
        Email: lyuhe233@stu.scu.edu.cn,
        \{zenghuolin, wangjunren, yanghuazhen\}@wchscu.cn, \\
        \{hlc, yongchen, lizhirui\}@stu.scu.edu.cn, 
        andreas.maier@fau.de,
        siming.bayer@fau.de,
        songhuan@wchscu.cn
    }
}

\maketitle

\maketitle

\begingroup
\small
\noindent
\textcopyright~2026 IEEE. Personal use of this material is permitted.
Permission from IEEE must be obtained for all other uses, including
reprinting/republishing this material for advertising or promotional purposes,
creating new collective works, resale or redistribution, or reuse of any
copyrighted component of this work in other works.

\vspace{0.5em}
\noindent
Accepted for publication in the 2026 48th Annual International Conference
of the IEEE Engineering in Medicine \& Biology Society (EMBC).
The published version will be available in IEEE Xplore.
\par
\endgroup
\vspace{0.5em}

\begin{abstract}
Real-world clinical data is inherently multimodal, providing complementary evidence that mirrors the practical necessity of jointly assessing multiple related outcomes. Although multi-task learning can improve efficiency by sharing information across outcomes, existing approaches often fail to balance shared representation learning with outcome-specific modeling. Hard parameter sharing can trigger negative transfer when task gradients conflict, while flexible sharing may still entangle shared and task-specific signals. To address this, we propose a multi-task framework built on a unified Transformer for multimodal fusion, augmented with Orthogonal Task Decomposition (OrthTD) to split patient representations into shared and task-specific subspaces and impose a geometric orthogonality constraint to reduce redundancy and isolate task-specific signals. We evaluated OrthTD on a real-world cohort of 12,430 surgical patients for predicting four outcomes. OrthTD achieved average AUC (area under the receiver operating characteristic curve) of 87.5\% and average AUPRC (area under the precision-recall curve) of 37.2\%, consistently outperformed advanced tabular and multi-task methods. Notably, OrthTD achieves substantial gains in AUPRC, indicating superior performance in identifying rare events within imbalanced clinical data. These results suggest that enforcing non-redundant shared and task-specific representations can improve multi-outcome prediction from multimodal clinical data.

\end{abstract}


\section{Introduction}
The rapid digitization of healthcare has generated massive amounts of real-world clinical data, creating new opportunities for clinical prediction models that support medical decision-making \cite{Ngiam2019}. Such data are heterogeneous, combining structured variables (demographics, comorbidities, laboratory tests, medications, procedures) with unstructured narratives in EMRs, as well as other data sources depending on the setting (e.g., imaging, waveforms) \cite{Garriga2023}. While these sources describe the same patient, they vary in information granularity, clinical perspective, and representational modality. A central problem is how to learn patient representations that can use this heterogeneity effectively. Artificial intelligence (AI) based representation learning provides a practical approach by integrating multi-source inputs and learning patient-level features directly from routine clinical records, reducing reliance on manual feature engineering \cite{Varghese2024}.

\begin{figure*}[!t]
\centering
\includegraphics[width=\textwidth]{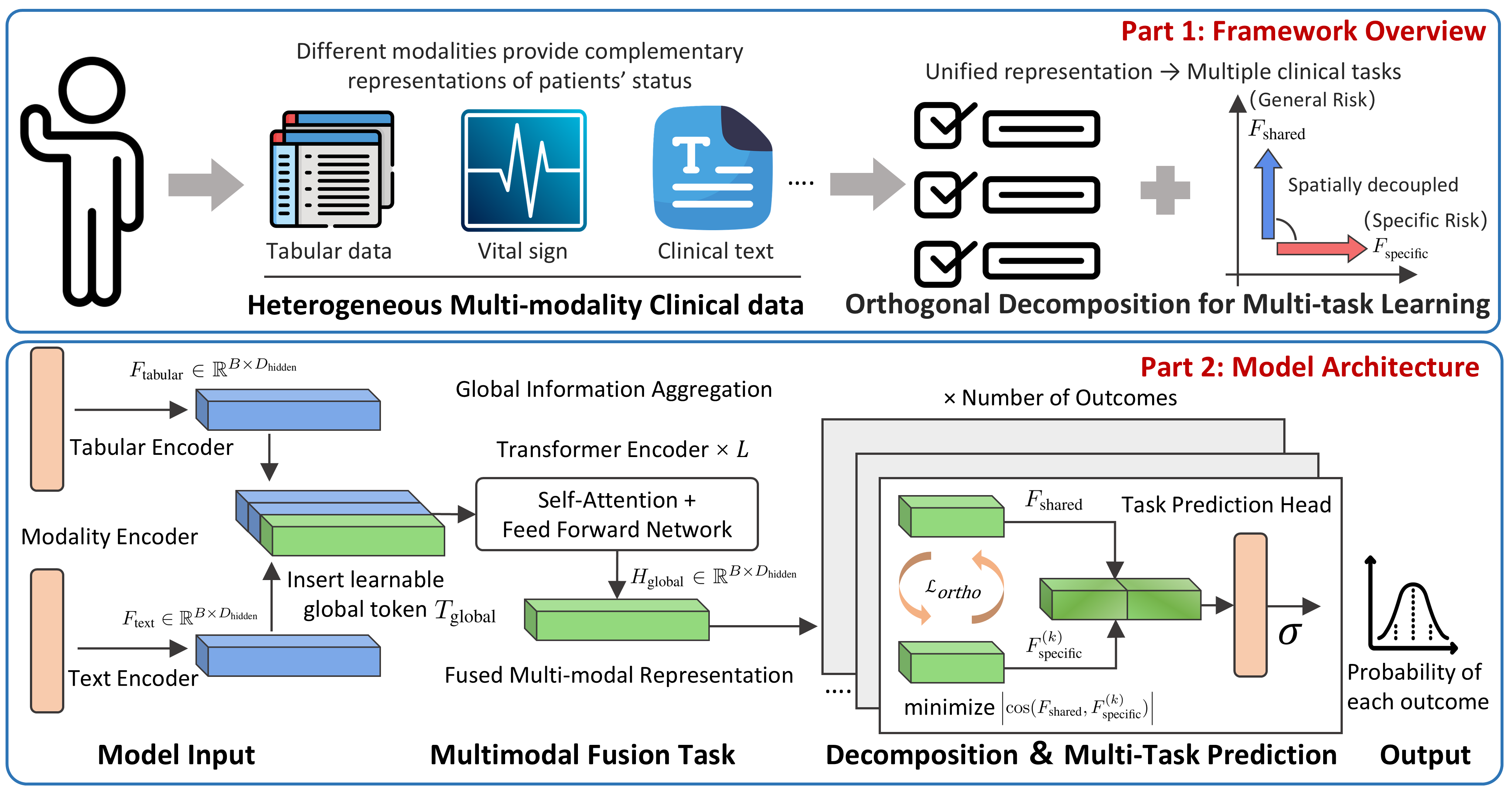}
\caption{Overview of the Orthogonal Task Decomposition (OrthTD) framework.
The figure is composed of two parts. Part 1 (Framework Overview) illustrates how heterogeneous multimodal clinical data (e.g., tabular variables, vital signs, and clinical text) are fused into a unified patient representation, which is then decomposed into a shared component capturing general risk and task-specific components encoding outcome-dependent risk, enabling structured multi-task learning. Part 2 (Model Architecture) details the implementation: modality-specific encoders project inputs into a common latent space, which is fused by a Transformer with a learnable global token to form a multimodal representation. This representation is then decomposed into shared and task-specific subspaces for each outcome, with an orthogonality regularization enforcing non-redundancy between them, and task-specific prediction heads produce probabilistic estimates for multiple clinical outcomes.}
\label{fig1}
\end{figure*}

Clinical decision-making often involves multiple related prediction tasks rather than a single outcome \cite{yang2024multi}. These tasks share part of the underlying patient factors, yet each task also depends on distinct cues. This naturally calls for learning clinical task representations that can capture both shared and task-specific information \cite{Xue2025}. Multi-task learning can improve data efficiency and generalization by enabling information sharing across tasks, but effective sharing is difficult in real-world clinical settings. Hard parameter sharing can cause negative transfer when task gradients conflict, and more flexible sharing strategies may still mix shared and task-specific signals within the same latent space. As a consequence, “shared” and task-specific representations can become redundant, with overlapping information that weakens specialization and destabilizes training. This challenge is further amplified when tasks have different label distributions and noise levels, where limited task-specific supervision makes it easier for shared patterns to dominate.

Related work spans two closely connected directions: multimodal clinical representation learning and multi-task learning. For multimodal learning, a common pipeline encodes structured variables and unstructured text separately (and other modalities when available) and then fuses them by feature concatenation, late fusion of logits, cross-attention, or unified tokenization followed by an encoder. These approaches aim to combine complementary information across modalities, but they often produce a single fused representation that is reused for all downstream tasks \cite{stahlschmidt2022multimodal}. For multi-task learning \cite{9392366}, widely used architectures include independent task-specific models, hard sharing with a shared backbone and multiple heads, and soft-sharing variants that adjust sharing strength through learnable feature mixing (e.g., cross-stitch \cite{MisraIshanandShrivastavaAbhinavandGuptaAbhinavandHebert2016}) or expert-routing (e.g., mixture-of-experts with task-specific gates \cite{Ma2018}). Another line of work focuses on optimization and balancing, such as loss reweighting (including uncertainty-based weighting \cite{Kendall_2018_CVPR}) or gradient-based conflict handling. Despite these advances, many existing methods do not directly control what information should be shared versus reserved for each task. In practice, even when separate branches or experts are introduced, shared and task-specific pathways can still encode overlapping signals, which reduces the value of decomposition and makes it harder to isolate task-specific cues, especially under different label distributions and limited positive supervision. Moreover, most multimodal and multi-task components are developed relatively independently, leaving the combined multimodal multi-task setting without a clear mechanism to prevent redundancy between shared and task-specific representations while keeping the fusion module modular. Our design is intended to explicitly separate shared and task-specific information after multimodal fusion and to discourage overlap between them.

To address these challenges, we propose OrthTD, a multimodal multi-task method that explicitly decomposes task representations and enforces a non-redundancy constraint. An overview of OrthTD is shown in Figure~\ref{fig1}. Our approach is based on the hypothesis that effective clinical modeling requires explicitly disentangling general patient risks from outcome-specific signals. OrthTD integrates tabular and textual data using a unified Transformer backbone and employs a Task Decomposition module to separate the latent representation into a shared subspace and multiple task-specific subspaces. Crucially, we introduce a geometric orthogonal constraint that forces these subspaces to be non-redundant. This ensures that shared features capture truly universal risk patterns, while task-specific features isolate unique signals required to distinguish individual complications. We validated our method from a large real-world clinical cohort. Our experiments in 12,430 surgical patients demonstrate that OrthTD effectively resolves the feature entanglement problem. The contributions of this work are as follows:
\begin{enumerate}
\item We propose a multimodal method for multi-task clinical prediction that jointly models structured tabular variables and unstructured clinical notes, validated on a prospective cohort of 12,430 patients.
\item We propose a task decomposition module that factorizes the fused patient representation into shared and task-specific subspaces.
\item We introduce an orthogonality-based regularization that discourages redundancy between shared and task-specific representations, and our method outperforms strong tabular, multimodal, and multi-task baselines in comparative experiments.
\end{enumerate}

\section{Materials and Methods}

\subsection{Data Source and Study Design}

This study used data from the China Surgery and Anesthesia Cohort (CSAC)\footnote{Additional details of CSAC are available at \url{https://biomedbdc.wchscu.cn/JoylabErasePM/joylab-portals-web/#/queue/01/index}}, an ongoing prospective multicenter cohort recruiting patients aged 40-65 years undergoing elective surgery under general anesthesia since July 15, 2020. The cohort profile has been described in detail previously \cite{Yang2024}. Briefly, cohort staff collected baseline data through preoperative interviews and ascertained postoperative outcomes via in-person (bedside) and remote (telephone/online) follow-up. Perioperative data were routinely extracted from hospital information systems, including the anesthesia information management system (AIMS) for intraoperative physiological data and the EMR system for comprehensive clinical care records. As of April 1, 2025, 18,709 of 19,884 screened patients were enrolled (baseline response rate: 94.09\%).

In this study, we selected postoperative complications as the multi-task prediction targets. Analyses were restricted to the primary center (West China Hospital, n = 12,430) from July 15, 2020 to April 1, 2025, which had fully integrated AIMS and EMR data enabling comprehensive multimodal modeling.  Postoperative complications occurring within seven days after surgery were defined according to the European Perioperative Clinical Outcome (EPCO) criteria \cite{Jammer2015}. The prediction outcomes included any EPCO complication (defined as the occurrence of any measured complications), postoperative pulmonary complications (PPCs), acute kidney injury (AKI), and unplanned ICU admission (ascertained from EMRs).

\subsection{Methods}
The model architecture is depicted in Figure~\ref{fig1}. Our approach builds a unified patient representation from heterogeneous clinical modalities and then structures this representation for multi-task prediction through explicit feature decomposition. A Transformer-based fusion module aggregates information across modalities into a global latent embedding, which is subsequently split into a shared subspace encoding overall risk and task-specific subspaces capturing outcome-dependent factors. By enforcing orthogonality between these components, the model is encouraged to separate common and outcome-specific signals, enabling both effective information sharing and robust specialization across tasks. In the following we will detail the core elements and the implementation of our method.
\subsubsection{Multimodal Data Fusion}

To convert raw multimodal inputs into unified feature representations, we tokenized all heterogeneous data \cite{Vaswani2017}. Categorical features were processed through embedding layers, concatenated with continuous numerical features, and projected to obtain the tabular modality representation \ensuremath{F_{\text{tabular}} \in \mathbb{R}^{B \times D_{\text{hidden}}}}. For unstructured text, we extracted the \ensuremath{[\text{CLS}]} token from a shared-weight BERT \cite{Devlin2019}, followed by linear projection to obtain \ensuremath{F_{\text{text}} \in \mathbb{R}^{B \times D_{\text{hidden}}}}. For vital sign time-series data with variable lengths due to different surgical durations, we converted them into prior features using medically predefined feature extraction methods and incorporated them into the tabular modality \cite{Wesselink2018}. After concatenating different modality features, a learnable global token \ensuremath{T_{\text{global}} \in \mathbb{R}^{D_{\text{hidden}}}} was inserted to aggregate global information from the entire input sequence, forming the comprehensive feature for downstream tasks. The sequence was then processed through \ensuremath{L} stacked Transformer Encoders. The output corresponding to the global token position, denoted as \ensuremath{H_{\text{global}} \in \mathbb{R}^{B \times D_{\text{hidden}}}}, serves as the fused multi-modal representation.

\subsubsection{Task Decomposition}

In multi-task learning, different clinical outcomes often share common risk factors while also having task-specific predictors. To explicitly model this characteristic, we propose a Task Decomposition module that separates the fused representation into shared and task-specific components. Given the fused representation \ensuremath{H_{\text{global}}}, we first extract shared features that are common across all prediction tasks:
\begin{equation}
F_{\text{shared}} = \text{LayerNorm}(\text{GELU}(W_{\text{shared}} \cdot H_{\text{global}} + b_{\text{shared}})), \label{eq1}
\end{equation}
where \ensuremath{W_{\text{shared}} \in \mathbb{R}^{D_{\text{hidden}} \times D_{\text{shared}}}}. The shared features capture general patient characteristics that influence multiple outcomes. For each task \ensuremath{k \in \{1, 2, ..., K\}}, we learn a separate projection to extract task-specific signals:
\begin{equation}
F_{\text{specific}}^{(k)} = \text{LayerNorm}(\text{GELU}(W_{\text{specific}}^{(k)} \cdot H_{\text{global}} + b_{\text{specific}}^{(k)})), \label{eq2}
\end{equation}
where \ensuremath{W_{\text{specific}}^{(k)} \in \mathbb{R}^{D_{\text{hidden}} \times D_{\text{specific}}}}. Each task-specific projection learns to extract features uniquely relevant to that particular outcome. We concatenate the shared and specific features to form the final representation for each task \ensuremath{F_{\text{task}}^{(k)}}.

This decomposition provides an explicit inductive bias: the model is encouraged to learn features that are either universally predictive (shared) or selectively predictive (task-specific), rather than learning entangled representations that conflate these two types of information. Finally, each task feature \ensuremath{F_{\text{task}}^{(k)}} is sent into a task-specific prediction head (MLP), which outputs the probability of complication through a sigmoid activation.

\subsubsection{Orthogonal Constraint}

While the Task Decomposition architecture provides explicit separation between shared and task-specific representations, there is no guarantee that these learned features will be non-redundant. Without proper constraints, the shared features and task-specific features may encode similar information, limiting the effectiveness of the decomposition.

To address this issue, we introduce an orthogonal constraint that encourages \ensuremath{F_{\text{shared}}} and \ensuremath{F_{\text{specific}}^{(k)}} to be geometrically orthogonal, thereby ensuring that they capture complementary rather than redundant information. We define the orthogonal regularization term as the average absolute cosine similarity between shared and task-specific features:
\begin{equation}
\mathcal{L}_{\text{ortho}} = \frac{1}{K} \sum_{k=1}^{K} \left| \cos(F_{\text{shared}}, F_{\text{specific}}^{(k)}) \right|. \label{eq3}
\end{equation}
When this loss approaches zero, the shared and task-specific features become orthogonal in the representation space, indicating that they encode distinct types of information. The final loss function combines the task-specific prediction losses with the orthogonal regularization:
\begin{equation}
\mathcal{L} = \mathcal{L}_{\text{task}} + \lambda_{\text{ortho}} \cdot \mathcal{L}_{\text{ortho}}, \label{eq4}
\end{equation}
where \ensuremath{\mathcal{L}_{\text{task}}} is the multi-task classification loss (asymmetric loss \cite{Ridnik2021} for handling class imbalance), and \ensuremath{\lambda_{\text{ortho}}} controls the strength of the orthogonal constraint.

\subsubsection{Implementation details}

All experiments were conducted on a workstation running Ubuntu 22.04 with an Intel Xeon Gold 6530 CPU and two NVIDIA RTX 5090 GPUs. The model was implemented in Python 3.12 using PyTorch 2.7 \cite{Paszke2019} and CUDA 12.8. We used BERT-base-Chinese (hidden size 768) for text encoding and fine-tuned only its last Transformer layer while freezing the remaining layers. All modality features were projected to a unified hidden dimension \ensuremath{D_{\text{hidden}} = 240}. The multimodal fusion backbone comprised 4 Transformer encoder layers with 8 attention heads each. Training was performed for 40 epochs using AdamW \cite{Loshchilov}, with learning rates of 1e-4 for the main model and 1e-5 for BERT, a cosine learning-rate schedule with 10\% warm-up (4 epochs), and a batch size of 128. In the Task Decomposition module, we set the shared ratio $0.5$, assigning equal dimensions to the shared and task-specific subspaces. Regarding the orthogonal regularization weight, we observed that the model performance remained consistent for $\lambda_{ortho} \in [0.1, 0.3]$, while exceeding $0.3$ led to training instability. For simplicity, we fixed $\lambda_{ortho}=0.1$ in all reported experiments. We hypothesize that optimal $D_{hidden}$ and $\lambda_{ortho}$ are likely correlated with modality count, data complexity, and the degree of task relatedness, and leave a systematic analysis to future work.

\begin{table}[htbp]
\caption{Incidence of Outcomes in the Train and Test Sets}
\centering
\begin{tabular}{lcc}
\toprule
\textbf{Outcome} & \textbf{Train set, n (\%)} & \textbf{Test set, n (\%)} \\
\midrule
Any EPCO complication & 1115 (12.8\%) & 471 (12.6\%) \\
PPCs & 949 (10.9\%) & 386 (10.4\%) \\
AKI & 77 (0.9\%) & 54 (1.4\%) \\
Unplanned ICU admission & 128 (1.5\%) & 42 (1.1\%) \\
\bottomrule
\end{tabular}
\label{tab1}
\end{table}

\section{Experiments and Results}

We conducted experiments to assess the discriminative ability and calibration of OrthTD, and to understand the contribution of each component. The cohort was randomly partitioned into train and test sets at a 70:30 ratio. The incidence of outcomes is summarized in Table~\ref{tab1}. We evaluated the proposed OrthTD model against comparative methods and conducted ablation studies. Three evaluation metrics were used: AUC and AUPRC (both reported in \%) for discrimination, and the Brier score for calibration.

Figure~\ref{fig1} shows the ROC and PR curves for all four prediction tasks. The model achieved strong discrimination for any EPCO complication and PPCs, which had higher incidence rates. For imbalanced outcomes (AKI and unplanned ICU admission), the model maintained high discrimination but with lower AUPRC values as expected for rare events.

\begin{figure}[!htbp]
\centering
\includegraphics[width=\columnwidth]{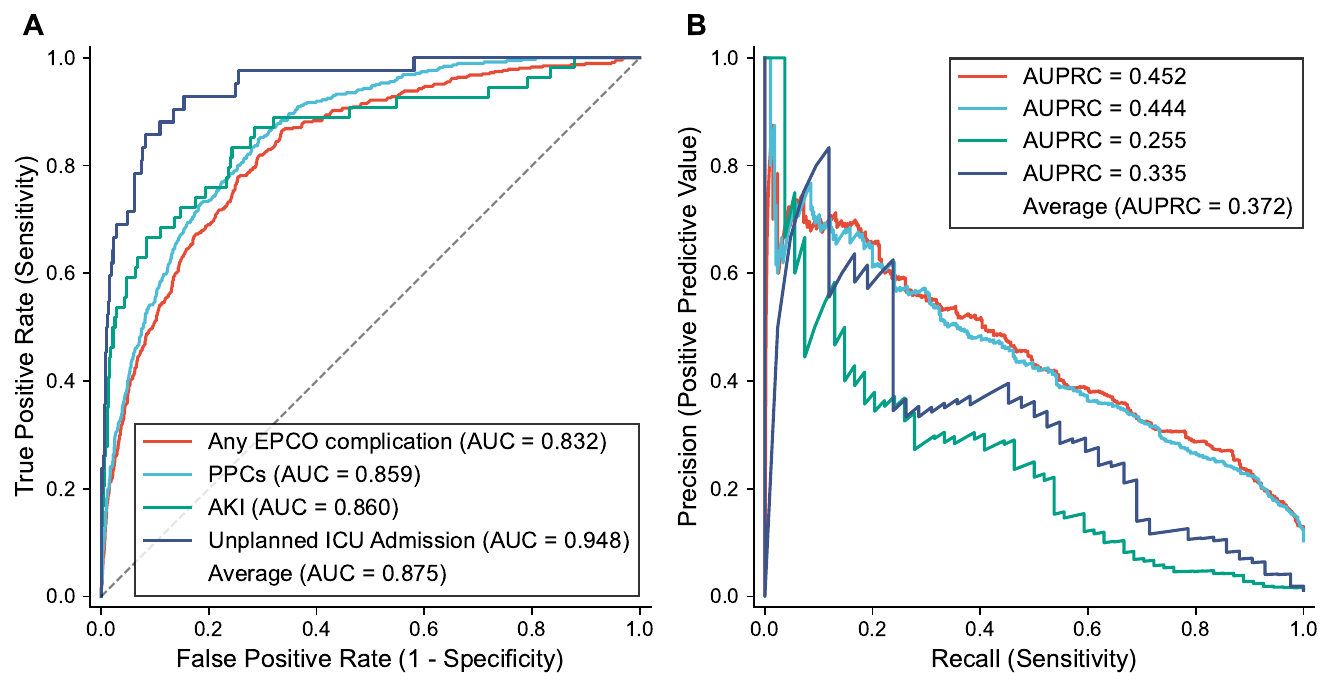}
\caption{Detailed performance of the proposed model.}
\label{fig2}
\end{figure}

\subsubsection{Comparative Analysis with Advanced Methods}

Table~\ref{tab:table2} shows the performance comparison between OrthTD and competing methods. We selected these baselines to cover strong tabular learning approaches from both classical machine learning and modern neural models, and to include representative multimodal fusion baselines for a fair comparison with OrthTD. Among classical machine learning methods, LightGBM \cite{NIPS2017_6449f44a} delivered the strongest performance, outperforming XGBoost \cite{Chen2016}, and serves as a strong machine learning baseline for tabular data. For advanced modern neural network tabular methods, TabPFN \cite{Hollmann2025} is a state-of-the-art (SOTA) baseline and achieved the best performance among unimodal models, while FT-Transformer \cite{NEURIPS2021_9d86d83f} is a strong attention-based tabular model. Multimodal baselines based on simple concatenation showed limited gains. OrthTD achieved the best overall performance, with a clear improvement on AUPRC, indicating better identification of positive cases under class imbalance.

\begin{table}[htbp]
\centering
\caption{Model Performance Comparison.}
\label{tab:table2}
\begin{tabular}{l l c c}
\toprule
\textbf{Model} & \textbf{Modality (Fusion)} & \textbf{AUC (\%)} & \textbf{AUPRC (\%)} \\
\midrule
XGBoost          & Tabular              & 85.5 & 29.7 \\
LightGBM         & Tabular              & 86.3 & 31.4 \\
FT-Transformer   & Tabular              & 86.7 & 32.5 \\
TabPFN           & Tabular              & 87.1 & 32.9 \\
MLP              & Tabular + Text (Concat) & 86.2 & 30.8 \\
Transformer      & Tabular + Text (Concat) & 84.7 & 28.6 \\
Ours (OrthTD)    & Tabular + Text (Fusion)  & \textbf{87.5} & \textbf{37.2} \\
\bottomrule
\end{tabular}
\end{table}

\subsubsection{Multi-Task Learning Strategy Comparison}

We compared OrthTD against five representative multi-task learning strategies that cover common design choices for feature sharing and task balancing, including independent training of proposed model, hard sharing, loss reweighting, feature mixing, and expert routing (Table~\ref{tab:table3}). Single-task learning with our models achieved AUC 86.9\% and AUPRC 34.5\%, avoiding negative transfer but failing to use shared patterns. Hard parameter sharing obtained AUC 85.8\% and AUPRC 33.7\%, with performance drop suggesting that forced representation sharing introduces conflicting gradients. Uncertainty weighting \cite{Kendall_2018_CVPR} improved performance to AUC 86.7\% and AUPRC 35.6\%, showing the importance of task balancing. Cross-stitch networks \cite{MisraIshanandShrivastavaAbhinavandGuptaAbhinavandHebert2016} achieved AUC 87.1\% and AUPRC 36.1\% through flexible feature sharing. Multi-gate mixture-of-experts (MMoE) \cite{Ma2018} yielded AUC 87.3\% and AUPRC 35.8\%, while its lower AUPRC suggests that explicit orthogonal constraints may provide a stronger inductive bias under our setting. Our OrthTD achieved the highest performance, suggesting that enforcing orthogonality between shared and task-specific features can mitigate negative transfer while preserving knowledge sharing.

\begin{table}[htbp]
\centering
\caption{Comparison of multi-task learning strategies.}
\label{tab:table3}
\begin{tabular}{l c c c}
\toprule
\textbf{Method} & \textbf{Type} & \textbf{AUC (\%)} & \textbf{AUPRC (\%)} \\
\midrule
Ours                         & MTL         & \textbf{87.5} & \textbf{37.2} \\
Ours (Single-Task)           & Independent & 86.9 & 34.5 \\
Hard Parameter Sharing       & MTL         & 85.8 & 33.7 \\
Uncertainty Weighting        & MTL         & 86.7 & 35.6 \\
Cross-Stitch Networks        & MTL         & 87.1 & 36.1 \\
Multi-gate Mixture-of-Experts& MTL         & 87.3 & 35.8 \\
\bottomrule
\end{tabular}
\end{table}

\subsubsection{Ablation Study and Detail Model Performance}

\begin{figure}[!htbp]
\centering
\includegraphics[width=\columnwidth]{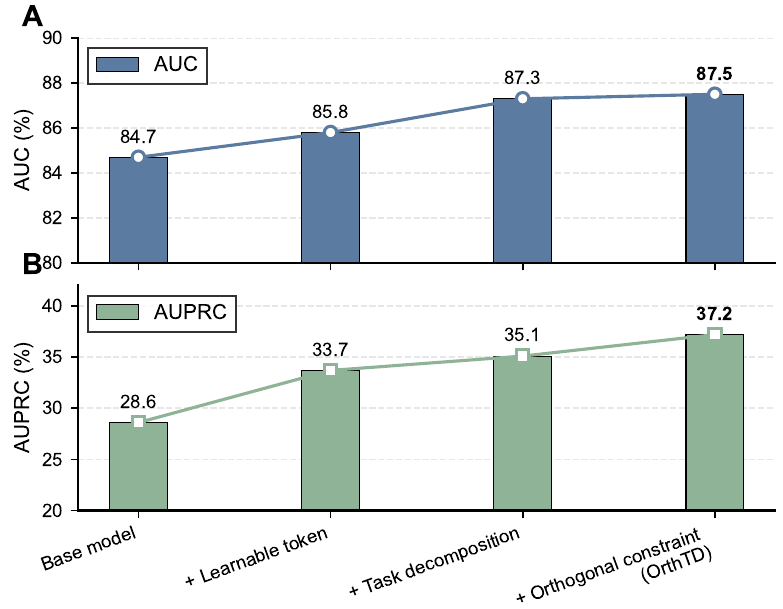}
\caption{Performance in the ablation study of the proposed model.}
\label{fig3}
\end{figure}

Figure~\ref{fig3} shows the ablation study results. Starting from the vanilla Transformer (base model), adding a learnable global token under hard sharing improved performance to AUC 85.8\%, AUPRC 33.7\%. The 5.1-point AUPRC gain shows that explicit global information aggregation is important for multimodal fusion. Adding task decomposition further improved performance (AUC 87.3\%, AUPRC 35.1\%), validating that explicitly modeling shared and task-specific risk factors provides useful inductive bias. Finally, enforcing orthogonality between shared and task-specific features achieved the best performance. The AUPRC improvement confirms that orthogonality prevents redundant information and enhances feature separation. The final model achieved a Brier score of 0.077, indicating reasonably calibration.

\section{Discussion}
Integrating heterogeneous multimodal data is fundamental for clinical risk prediction, yet effectively balancing shared knowledge with task-specific signals remains a critical challenge in multi-task learning. Enforcing geometric orthogonality disentangles shared and task-specific representations, mitigating negative transfer and improving discrimination under both imbalanced and extremely imbalanced outcome settings. Our approach achieves consistent SOTA performance across all experiments and metrics. This pattern is important because several targets in our setting are extremely rare, where improvements in precision–recall performance better reflect practical screening value than AUC alone. The ablation results further show that the performance gains are cumulative, indicating that multimodal fusion, task decomposition, and the orthogonal constraint each contribute to the final model behavior rather than serving as a single isolated trick.

A key challenge in multi-task learning for clinical outcomes is the coexistence of shared and task-specific risk factors. Hard parameter sharing can force incompatible gradients into the same representation and lead to negative transfer, while purely soft-sharing approaches may still mix signals that should remain distinct. OrthTD addresses this by explicitly decomposing the fused patient representation into shared features and task-specific features for each outcome. On top of this structure, the orthogonal constraint encourages these components to be non-redundant, promoting complementary information rather than duplicated content. The observed improvements over alternative multi-task strategies are consistent with this motivation, suggesting that separating general risk patterns from outcome-specific cues can improve overall performance.

The results also highlight the value of multimodal modeling for clinical prediction. Clinical narratives in EMRs often contain rich information that is not fully represented by fixed fields, such as nuanced comorbidity descriptions, symptom history, care plans, and clinician assessments. In modern medical practice, structured tabular data and EMR text are routinely available throughout the clinical workflow, making their combination a practical and widely applicable choice for risk modeling. Our fusion backbone uses a unified token-based representation and a learnable global token to aggregate information across modalities, which is supported by the clear gains when adding global aggregation in the ablation study. Together, these design choices provide a practical way to integrate heterogeneous medical data without requiring task-specific feature engineering for every new outcome.

OrthTD provides a generalized framework where task-specific heads and modality encoders can be expanded without altering the core decomposition mechanism. Future work will focus on scaling this architecture to broader clinical settings by incorporating extra data types, new prediction targets and advanced model components, such as sequence models for raw vital signs and domain-adapted language models to improve data efficiency. Additionally, we will strengthen real-world reliability by assessing subgroup robustness and stability under missing-modality scenarios.

\section{Conclusion}

In this work, we presented OrthTD, a multimodal multi-task learning method designed to model heterogeneous real-world clinical data. By explicitly decomposing representations and enforcing geometric orthogonality, our approach mitigates the trade-off between feature sharing and task specialization. The empirical results on a large-scale clinical cohort confirm that separating general patient risks from outcome-specific signals mitigates negative transfer and reduces feature redundancy. Consequently, OrthTD delivers superior discrimination compared to strong unimodal and multi-task baselines, particularly for detecting low-incidence events. Furthermore, the unified fusion backbone allows for the flexible integration of additional modalities, making the framework a scalable solution for comprehensive clinical risk monitoring.

\section*{Acknowledgment}

We thank all the team members and colleagues involved in CSAC for their support. The computations in this paper were supported by the High Performance Computing platform at West China Biomedical Big Data Center, West China Hospital, Sichuan University.

\bibliographystyle{IEEEtran}
\bibliography{refs}

\end{document}